\documentclass[a4paper, 10pt, conference]{ieeeconf}      
\usepackage{FG2021}
\let\labelindent\relax
\usepackage{enumitem}
\usepackage{adjustbox}
\usepackage{threeparttable}
\usepackage{multirow}

\FGfinalcopy 

\def\FGPaperID{352} 

\title{\LARGE \bf
Which CNNs and Training Settings to Choose for Action Unit Detection? A Study Based on a Large-Scale Dataset
}

\author{\parbox{16cm}{\centering
    {\large Mina Bishay, Ahmed Ghoneim, Mohamed Ashraf and Mohammad Mavadati}\\
    {\normalsize
    Smart Eye AB}}
}

\begin{document}

\ifFGfinal
\thispagestyle{empty}
\pagestyle{empty}
\else
\author{Anonymous FG2021 submission\\ Paper ID \FGPaperID \\}
\pagestyle{plain}
\fi
\maketitle

\begin{abstract}


In this paper we explore the influence of some frequently used Convolutional Neural Networks (CNNs), training settings, and training set structures, on Action Unit (AU) detection. Specifically, we first compare 10 different shallow and deep CNNs in AU detection. Second, we investigate how the different training settings (i.e. centering/normalizing the inputs, using different augmentation severities, and balancing the data) impact the performance in AU detection. Third, we explore the effect of increasing the number of labelled subjects and frames in the training set on the AU detection performance. These comparisons provide the research community with useful tips about the choice of different CNNs and training settings in AU detection. In our analysis, we use a large-scale naturalistic dataset, consisting of $\sim$55K videos captured in the wild. To the best of our knowledge, there is no work that had investigated the impact of such settings on a large-scale AU dataset.


\end{abstract}



\section{INTRODUCTION}

Understanding and recognition of facial expressions are critical to nonverbal communication. Motivated by a wide range of expression-related applications, the Computer Vision community has become increasingly interested in developing algorithms for AU detection in the last two decades \cite{martinez2017automatic, li2020deep}. Recently, researchers focused on learning deep features using CNNs, as they have shown to learn better features than conventional hand-crafted features \cite{le2011learning, krizhevsky2012imagenet, antipov2015learned}. Across the literature, different CNN architectures, training settings (e.g. input centering/normalization, augmentation severity, data balancing), and training set structures (e.g. different number of labelled frames and subjects) have been used in AU detection. However, to the best of our knowledge, it has not been explored how these different CNNs and training settings affect the AU detection performance.

\begin{figure*} [t!]
  \centering{\includegraphics[width=0.93 \linewidth]{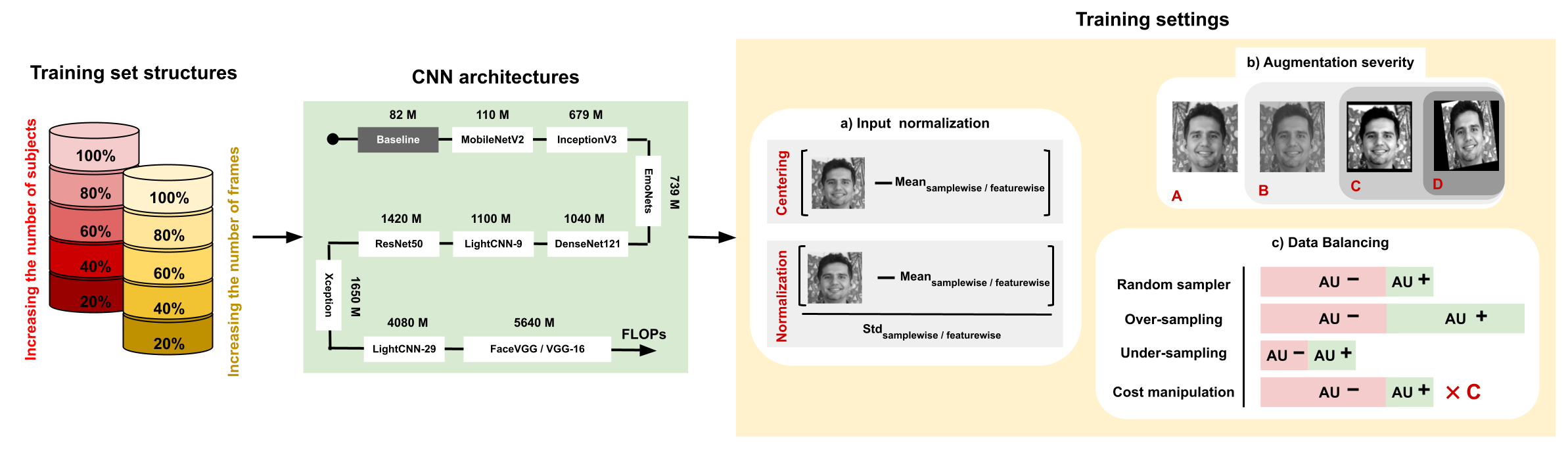}}
  \caption{Comparing different CNN architectures, training settings, and training set structures.}
  \label{fig:table_5}
  \label{fig_all}
  \vspace{-4mm}
\end{figure*}

The objective of this paper is to compare the impact of a variety of CNN architectures and training settings on AU detection performance. To have reliable and generalizable results, we use in our analysis a large-scale dataset consisting of around 55K videos (from 90+ countries) for participants watching commercial ads. Participants’ videos were captured at different recording conditions (i.e. in the wild), and with fully spontaneous facial expressions. This is different from other existing datasets with relatively limited subjects, recording conditions, and/or demographic diversity.  The videos have been annotated for different AUs by a group of trained Facial Action Coding System (FACS \cite{EkmanBook97}) coders. Our analysis includes comparing various a) CNN architectures, b) training settings, and c) training set structures (comparisons are shown in Fig. \ref{fig_all}).


\textbf{Different CNN architectures} have been used across the literature for AU detection. Some works used shallow CNNs for feature extraction \cite{ghosh2015multi, kahou2016emonets, tu2019idennet}, while others trained deeper CNNs with relatively more layers, filters, and parameters \cite{zhao2016deep, hayale2019facial}. Deeper CNNs increase the computational and memory cost of the whole AU detection pipeline, however, it is not clear how such deep CNNs perform in comparison to other shallow CNNs, as well as how those deep CNNs compare to each other. In this paper, we will compare 10 different CNN architectures in AU detection.



\textbf{Training settings} used for CNNs vary across the literature, and their choice impacts the performance. Some works used the raw face images for training CNNs, while others normalize or center the images before processing them. In addition, data augmentation is one of the training tricks for improving feature extraction, and in AU detection the augmentation severity varies from using just a few augmenters to using a more severe augmentation. Furthermore, the training data is always imbalanced, and needs to get balanced to avoid biasing the classifier to the most frequent class. Some researchers balance the data by oversampling or undersampling, while others manipulate the cost term associated with less frequent classes. In this work, we will investigate how the different training settings affect the AU detection performance. 


\textbf{Training set structure.} Different datasets have been used for training CNNs in AU detection -- those datasets have different number of frames and subjects. In this work, we will explore the impact of increasing the number of training subjects and frames on the AU detection performance. Results should give the community useful insights about future data collection and labelling in AU detection.




\section{Baseline Settings}

In this section we describe the dataset, CNN architecture, and experimental settings used for our baseline experiment.

\subsection{Dataset}


Many datasets have been captured and annotated for spontaneous AUs in the last ten years (e.g. DISFA \cite{mavadati2013disfa}, UNBC \cite{lucey2011painful}, BP4D \cite{zhang2014bp4d}), nonetheless, these datasets have relatively limited number of participants, recording conditions, and/or diversity in demographics. For our analysis, we use a large-scale dataset that was captured in the wild, and with fully spontaneous facial behaviour. This dataset has $\sim$55K videos of participants with diverse age, gender and ethnicity. Note that the resolution and frame rate varies across the dataset as different recording devices (e.g. laptops, mobile phones) have been used, and also the video length varies depending on the ad length. For the experiments, we divide it into 40.9K videos for training, 5.9K for validation, and 8.2K for testing.



The web-based approach described in \cite{mcduff2013affectiva, mcduff2018fed+} enabled us to collect thousands of videos for participants watching commercial ads worldwide (from 90+ countries). The collected videos were annotated for the presence of AUs using trained FACS coders. In addition, the videos were labelled for gender  (55\% Female, 37\% Male, 8\% uncertain) and ethnicity (37\% Caucasian, 24\% East Asian, 14\% South Asian, 13\% Latin, 9\% African, 3\% uncertain). A part of this dataset was made available to the research community through AM-FED \cite{mcduff2013affectiva} and AM-FED+ \cite{mcduff2018fed+}. 



\subsection{Architecture}

\textbf{Preprocessing} consists of 3 main steps. First, we localize the participant’s face region by using a face detector trained in the wild. Then, we detect 4 landmarks (i.e. outer eye corners, nose tip and chin) on the face. We use these landmarks to align the eyes horizontally. Finally, the aligned faces are converted to grayscale, scaled to a fixed size of 96$\times$96, and passed as an input to a CNN. 




\textbf{A single CNN model} is jointly trained for the detection of 12 AUs (AUs are given in Table \ref{table1}). The CNN consists of 4 convolutional and 2 fully-connected layers. The 4 convolutional layers have 32, 32, 64, and 64 filters respectively. All the convolutional filters have a kernel of size 3$\times$3. A max-pooling layer with a filter of size 2$\times$2 is used after each convolutional layer. The activation function used in the convolutional layers is the ReLU function \cite{nair2010rectified}. The first fully-connected layer has 256 neurons, while the second has 12 sigmoid units representing the predictions of the 12 AUs.

%




\subsection{Experimental Settings}

\textbf{Training settings.} Most of the AUs in our dataset have a high ratio of negative to positive examples (i.e. highly imbalanced). In our analysis, we use oversampling to balance the data. Specifically, we sample 4 positive and 4 negative images for each AU, which results in 96 images (12 AUs $\times$ 8 images) for each training batch. The positive and negative examples are selected randomly from the training set videos. We use Binary Cross Entropy (BCE) for calculating the loss, and the total loss is the average of the BCE losses calculated over each AU images. 






In all the experiments, the CNN is initialized randomly from the same seed, and is trained for 600 epochs using the Adagrad optimizer with an initial learning rate set to 0.005. For each epoch, 300 batches are sampled for training and 300 batches for validation.  The training batches are augmented by random flipping, shifting, zooming, etc. For testing, the best validation model is tested on 3000 batches sampled from the testing set. Accuracy is used for evaluating performance, as we have an equal number of positive and negative examples in the validation and testing batches. 



\begin{table*}[!t]
\caption{The accuracy and computational complexity across different CNN architectures.}
\label{table1}
\begin{adjustbox}{width=0.99 \textwidth,center}
\begin{threeparttable}
    \begin{tabular}{|c||c||c|c||c||c||c||c||c|c|c|c|c|}
 \hline
 \textbf{Exp.} & \textbf{Baseline} & \textbf{MobileNetV2 \cite{sandler2018mobilenetv2}} & \textbf{InceptionV3 \cite{szegedy2016rethinking}} & \textbf{EmoNets \cite{kahou2016emonets}} & \textbf{DenseNet121 \cite{huang2017densely}} & \textbf{LightCNN-9 \cite{wu2018light}} & \textbf{ResNet50 \cite{he2016deep}} & \textbf{Xception \cite{chollet2017xception}} & \textbf{LightCNN-29 \cite{wu2018light}} & \textbf{FaceVGG \cite{parkhi2015deep}} & \textbf{VGG-16 \cite{simonyan2014very}}  \\
 
 \hline
AU01 & 75.5 & 69.8 & 76.8 & 77.3 & 72.5 & 75.4 & \textbf{82.8} & 75.9 & 80.7 & 79.9 & [82.0]   \\ \hline
AU02 & 67.1 & 68.0 & 65.0 & 69.0 & 70.0 & 70.6 & 69.2 & 69.4 & [70.8] & 70.0 & \textbf{71.2}   \\ \hline
AU04 & 68.5 & 74.0 & 74.0 & 72.3 & 73.0 & 76.1 & 70.6 & 75.6 & [79.7] & 75.7 & \textbf{80.2}   \\ \hline
AU06 & 83.5 & 83.9 & 85.4 & 85.2 & 83.7 & 85.6 & 85.1 & 84.7 & \textbf{86.1} & 85.4 & [85.9]   \\ \hline
AU07 & 69.3 & 68.6 & 65.7 & 69.1 & 67.5 & 72.2 & 73.1 & 70.5 & \textbf{76.6} & 72.9 & [75.5]   \\ \hline
AU09 & 77.2 & 72.0 & 75.1 & [77.9] & \textbf{79.7} & 73.1 & 74.3 & 75.8 & 73.0 & 77.0 & 75.6   \\ \hline
AU12 & 84.8 & 82.0 & 84.1 & 86.4 & 80.9 & 86.9 & 82.8 & 83.5 & \textbf{87.6} & 86.9 & [87.1]   \\ \hline
AU15 & 78.6 & 80.0 & 75.1 & 80.2 & \textbf{84.1} & 82.2 & 82.3 & 79.5 & 82.2 & 81.3 & [83.9]   \\ \hline
AU17 & 76.9 & 73.2 & 75.1 & 77.9 & 79.3 & 80.9 & 80.2 & 76.3 & [81.3] & 80.8 & \textbf{82.6}   \\ \hline
AU24 & 77.6 & 70.2 & 72.0 & 76.6 & 74.5 & 79.8 & 73.0 & 70.3 & [82.1] & 80.0 & \textbf{84.2}   \\ \hline
AU25 & 72.7 & 77.2 & 77.1 & 77.4 & 78.8 & 80.3 & 79.7 & 73.2 & [80.6] & 80.5 & \textbf{83.9}   \\ \hline
AU28 & 83.8 & 85.4 & 85.6 & 84.9 & 84.5 & 86.8 & 86.9 & 86.2 & [87.2] & 86.5 & \textbf{89.5}   \\ \hline \hline
\textbf{AVG} & 76.3 & 75.4 & 75.9 & 77.8 & 77.4 & 79.1 & 78.3 & 76.7 & [80.6] & 79.7 & \textbf{81.8}   \\ \hline

 \hline
 \hline
 FLOPs & 81.6M & 110.1M & 679.2M & 739.4M & 1040M & 1100M & 1420M & 1650M & 4080M & 5640M & 5640M \\

 \hline
 \#params & 659K &	2,276K & 21,831K & 567K & 7,052K & 3,649K &	23,587K & 20,890K & 10,741K & 14,721K & 14,721K \\
 \hline
 Memory & 4.71MB &	16.44MB & 91.41MB &	6.32MB & 49.15MB & 31.76MB & 102.95MB &	101.07MB & 72.22MB & 69.97MB & 69.97MB \\
 \hline

 \hline
    \end{tabular}
\begin{tablenotes}
\item Bold numbers indicate the best performance, while bracketed numbers indicate the second best.
\end{tablenotes}
\end{threeparttable}
\end{adjustbox}
\vspace{-4mm}
\end{table*}


\section{CNN Architectures}

Features learned by Deep Neural Networks have shown better performance than hand-crafted ones across different applications \cite{krizhevsky2012imagenet, le2011learning, antipov2015learned}.  Subsequently, appearance features learned by CNNs have been immensely used in facial expression analysis in the last few years \cite{martinez2017automatic, li2020deep}. Some works used relatively shallow CNNs (with 3-5 convolutional layers) \cite{ghosh2015multi, jaiswal2016deep, kahou2016emonets, bishay2017fusing, tu2019idennet}, while others used deeper CNNs, that have been trained on other classification tasks like object or face recognition \cite{zhao2016deep, bishay2019schinet, hayale2019facial, vemulapalli2019compact}. The computational and memory cost of the AU detection pipeline increases notably as the CNN complexity increases. However, it is not clearly explored how the different shallow and deep CNNs affect the performance in AU detection. 




In this section, we clarify the effect of using 10 different CNNs in AU detection, so other researchers can properly consider deploying the best ones in their modeling, according to the available computational power. Some of these CNNs are shallow like the baseline CNN, the EmoNets CNN \cite{kahou2016emonets}, and the LightCNN-9 \cite{wu2018light}, while others are deep like VGG16 \cite{simonyan2014very}, ResNet50 \cite{he2016deep}, and LightCNN-29 \cite{wu2018light}. The 10 CNNs vary a lot in the number of parameters, CNN layers, FLOPs, and memory size, as shown in Table \ref{table1}. For all the CNNs (except the baseline and EmoNets CNNs), we first replace the last fully-connected layer with another layer having a number of neurons equal to the predicted AUs. Then, we initialize the CNN using pre-trained models on object/face recognition. Finally, we fine-tune the CNNs using an Adam optimiser with 0.0001 learning rate.



Table \ref{table1} shows the performance across the different CNNs. From Table \ref{table1}, we can first conclude that VGG-16 and LightCNN-29 are performing the best and the second best on average. VGG-16 and LightCNN-29 perform either the best or the second best across 11 and 9 AUs (out of 12 AUs), respectively. Second, MobileNetV2 and InceptionV3 are the worst performing networks across the compared CNNs. Third, although VGG-16 and Face VGG have the same structure, and only varying in the classification task used for pre-training the CNN, VGG-16 has better accuracy than the Face VGG -- this is aligned with the findings in \cite{niinuma2019unmasking}. Note that \cite{niinuma2019unmasking} has a relatively smaller dataset, and the comparison is only limited to VGG-16 and Face VGG. Finally, LightCNN-9 is a good choice for applications with a limited computational power, as it has relatively good performance and low computational cost, compared to the best-performing CNNs (VGG-16 and LightCNN-29).

\section{Training Settings}

\subsection{Input Centering/Normalization}

Some works in AU detection used the raw face images as input to the CNNs \cite{jung2015deep, zhao2016deep, bishay2017fusing, chu2017learning, bishay2019schinet, tu2019idennet}, while others centered (i.e. subtract the mean) or normalized the images (i.e. subtract the mean and divide by standard deviation) \cite{jaiswal2016deep, hayale2019facial, gudi2015deep} before passing them to the CNNs. For \textit{Samplewise centering/normalization}, the mean and standard deviation are calculated over each sample/frame, while for \textit{Featurewise centering/normalization} the mean and standard deviation are calculated over the whole training set. Centering and normalization helps in improving the training process. 


In this section, we compare different input normalization/centering methods for training the baseline CNN. Table \ref{table_all} shows the performance achieved by using the raw, centered, and normalized inputs. From Table \ref{table_all}, we can see that using the samplewise centering/normalization and the featurewise normalization improves the performance by $\sim$1\%, and is recommended for future work, while the featurewise centering has close performance to using the raw inputs.

\begin{table*}[ht]
\caption{The accuracy across different input normalization, data augmentation, and data balancing techniques.}
\label{table_all}
\begin{adjustbox}{width=0.99 \textwidth, center}
\begin{threeparttable}
  \centering
    \begin{tabular}{|c||c|c|c|c|c||c|c|c|c||c|c|c|c|}
 \hline

 &   \multicolumn{5}{|c|}{\textbf{Input centering/normalization}} & \multicolumn{4}{|c|}{\textbf{Severity of augmentation}}  & \multicolumn{4}{|c|}{\textbf{Data Balancing}}   \\\cline{2-14}
\textbf{Exp.} & \textbf{Raw} & \textbf{Samplewise}  & \textbf{Featurewise}  & \textbf{Samplewise}  & \textbf{Featurewise} & \textbf{Group A} & \textbf{Group B} & \textbf{Group C} & \textbf{Group D} & \textbf{Random} & \textbf{Under-} & \textbf{Over-} & \textbf{Cost}  \\
 & \textbf{input} &  \textbf{Centering} &  \textbf{Centering} &  \textbf{Normalization} &  \textbf{Normalization} & \textbf{augmenters} & \textbf{augmenters} & \textbf{augmenters} & \textbf{augmenters} & \textbf{sampling} & \textbf{sampling} & \textbf{sampling} & \textbf{manipulation}  \\\hline

AU01 & 75.5 & 76.0 & 73.9 & 74.5 & 73.2 & 65.2 & 74.5 & 71.4 & 77.7 & 72.6 & 72.7 & 75.5 & 75.4    \\ \hline
AU02 & 67.1 & 67.8 & 67.9 & 68.2 & 68.8 & 67.3 & 67.4 & 68.0 & 69.1 & 69.6 & 66.4 & 67.1 & 68.8   \\ \hline
AU04 & 68.5 & 72.5 & 67.7 & 70.3 & 70.4 & 68.4 & 70.3 & 73.0 & 72.2 & 73.9 & 67.9 & 68.5 & 71.3   \\ \hline
AU06 & 83.5 & 84.7 & 83.4 & 84.0 & 83.3 & 82.3 & 83.4 & 83.9 & 83.5 & 81.9 & 83.2 & 83.5 & 83.1 \\ \hline
AU07 & 69.3 & 70.5 & 68.7 & 69.9 & 70.5 & 63.5 & 68.6 & 71.7 & 72.2 & 71.8 & 69.4 & 69.3 & 69.1   \\ \hline
AU09 & 77.2 & 76.0 & 77.2 & 77.2 & 78.9 & 73.5 & 75.1 & 77.8 & 76.6 & 73.5 & 76.8 & 77.2 & 75.1  \\ \hline
AU15 & 78.6 & 79.9 & 79.0 & 80.1 & 80.5 & 77.0 & 77.5 & 79.9 & 79.1 & 75.6 & 75.6 & 78.6 & 79.5  \\ \hline
AU17 & 76.9 & 77.8 & 76.7 & 78.5 & 78.4 & 75.3 & 77.7 & 78.8 & 77.6 & 73.9 & 75.7 & 76.9 & 77.0  \\ \hline
AU24 & 77.6 & 77.6 & 77.6 & 78.0 & 78.3 & 75.8 & 75.4 & 77.1 & 78.4 & 73.8 & 75.6 & 77.6 & 75.9   \\ \hline
AU25 & 72.7 & 74.0 & 73.5 & 74.5 & 74.6  & 71.6 & 72.2 & 76.4 & 75.5 & 69.1 & 70.4 & 72.7 & 67.6   \\ \hline
AU28 & 83.8 & 85.1 & 83.6 & 85.4 & 84.7 & 82.6 & 83.9 & 86.0 & 84.5 & 77.5 & 82.4 & 83.8 & 83.6  \\ \hline
Smile & 84.8 & 85.4 & 85.0 & 85.6 & 84.9  & 83.5 & 85.5 & 85.8 & 84.8  & 85.0 & 83.8 & 84.8 & 85.8   \\ \hline  \hline
\textbf{AVG} & \textbf{76.3} & \textbf{77.3} & \textbf{76.2} & \textbf{77.2} & \textbf{77.2} & \textbf{73.8} & \textbf{76.0} & \textbf{77.5} & \textbf{77.6} & \textbf{74.8} & \textbf{75.0} & \textbf{76.3} & \textbf{76.0}   \\ \hline

    \end{tabular}
    \begin{tablenotes}
\item The baseline CNN is used for comparing the different training settings in this table.
\end{tablenotes}
\end{threeparttable}
\end{adjustbox}
\vspace{-4mm}
\end{table*}

\subsection{Severity of Augmentation}




Using data augmentation for training deep models improves the performance by reducing the overfitting problem \cite{shorten2019survey}. Different combinations of augmenters have been used across the literature for training the AU detection architectures, more specifically, \cite{bishay2017fusing, zhao2016deep, ghosh2015multi, shao2018deep, yang2019learning, lu2020self} used a few simple augmenters like flipping and cropping, while \cite{bishay2019schinet, hayale2019facial, lakshminarayana2020learning} used relatively more augmenters like flipping, shifting, and zooming, etc. It is not clear to what extent the variety of augmenters (i.e. the augmentation severity) impact the AU detection performance. In this section, we group several augmenters into 4 groups, ranging from no augmenters to a more various/severe combination of augmenters. Then, we compare how adding the simple to severe augmentation groups affects the performance of the baseline CNN. The augmentation groups are defined as follows:




\begin{itemize}[noitemsep] 
\item Group A includes no augmenters.
\item Group B includes horizontal flipping and changing randomly the brightness and contrast.
\item Group C includes Group B, and changing randomly the blurriness and zooming.
\item Group D includes Group C, horizontal and vertical shifting, and random rotation.
\end{itemize}




Table \ref{table_all} shows the results across the different augmenter groups. We can first conclude that although we are using a large-scale dataset, using data augmentation in the training helps improve the performance by $\sim$4\% (when comparing group A to D). Second, adding more augmenters (i.e. increasing the augmentation severity) improves the performance. Third, group C and D have very close performance, which means that at some point adding more augmenters will not add much to the performance, as the model has already learned a good feature representation. Subsequently, we recommend the research community to use a variety of different augmenters (similar to group C or D) for training their AU detection architectures. 

\subsection{Data Balancing}

Data balancing techniques have been used across different works, as the AU datasets used in the training are severely imbalanced (positive examples are typically limited), resulting in biasing the AU classifier towards the most frequent class. Oversampling and undersampling are two common techniques for balancing the data, by duplicating the positive examples, or removing negative examples in order to have an equal number of positive and negative examples \cite{chawla2009data}. In \cite{bishay2017fusing}, a method was proposed for balancing the data in a multilabel classifier by adjusting the loss term associated with each AU positive example, according to the ratio of the negative to positive examples in each training batch (called in our analysis the Cost manipulation method). In this section, we compare the random sampler to 3 balancing techniques (oversampling, undersampling, Cost manipulation) on our large-scale dataset. We use the baseline CNN in all our experiments. This will help the research community get insights about the improvement gained by using a balancing technique, as well as the best-performing technique. 

%




Table \ref{table_all} shows the results across different sampling techniques. From Table \ref{table_all}, we can first conclude that all the balancing techniques have better overall performance than the random sampler. Second, the oversampling technique has the best improvement (by 1.5\%), and the undersampling technique has the lowest improvement (by 0.2\%) across the balancing techniques -- this is because the undersampling technique discards a lot of the negative examples (i.e. reduces immensely the training set), while the oversampling method duplicates the positive frames without removing any frames. Third, the cost manipulation method has good performance, but slightly lower than the oversampling method. 
Subsequently, we recommend the research community to use the oversampling method for training AU detection architectures.
\section{Training Set Structure}

Across the different classification tasks (including AU detection), it has been commonly known that the more data we use for training CNNs, the better and more generalizable are the trained networks. The size of data used for training AU models has two main factors; a) the ``number of frames", and b) the ``number of subjects". It is not clear how increasing each of these factors affects the AU detection performance. 

In this section, we explore the impact of increasing the number of training subjects and frames on the performance. Specifically, we first fix the number of subjects (we use 100\% of the dataset subjects), and then train the model using 20\%, 40\%, 60\%, 80\%, and 100\% of the subjects' frames. Second, we fix the number of frames (we use 20\% of the total subjects' frames), and then train the model using 20\%, 40\%, 60\%, 80\%, and 100\% of the subjects. Note that each time we increase the number of subjects, we reduce the number of frames included from each subject by the same rate, to ensure that the total number of training frames is fixed. We use the baseline CNN for running all the experiments. This section should give researchers useful insights about future AU data collection and labelling.

%

\begin{figure} [t!]
  \centering{\includegraphics[width=0.99 \linewidth]{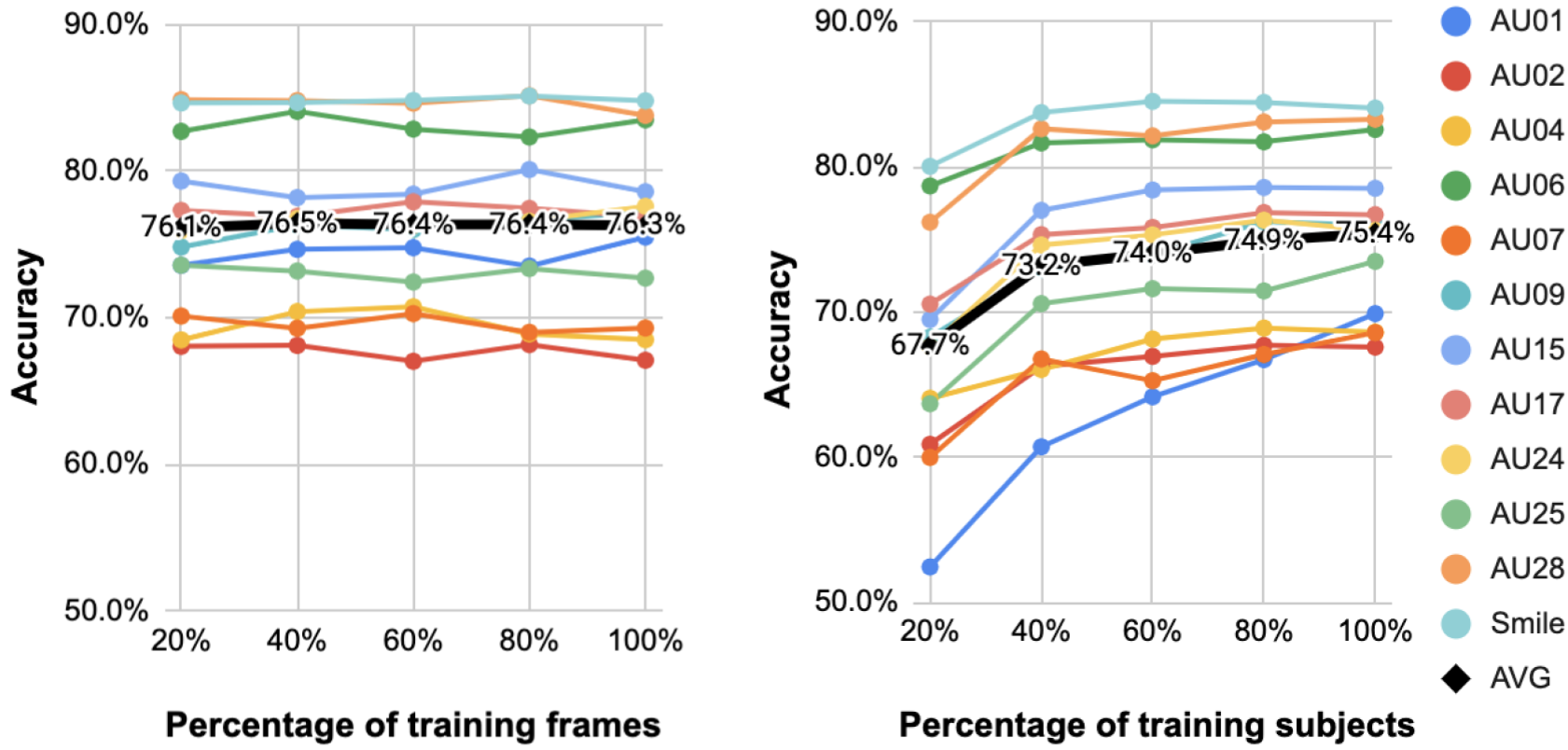}}
  \caption{The AU detection accuracy when using different number of frames (left) and subjects (right) for training the baseline CNN.}
  \label{fig2_train_set}
  \vspace{-4mm}
\end{figure}

Figure \ref{fig2_train_set} shows the results across the different number of frames and subjects. We can first conclude that the different number of frames have very close performance. Second, increasing the number of subjects in the training set improves the performance. Subsequently, increasing the number of labelled subjects in the training data is more important than increasing the number of labelled frames. This evidence informs us that it is better to put more effort into collecting more subjects and possibly labeling less frames for each subject. Ideally focusing on labeling mainly expressive subjects' frames, than labeling all the frames for less subjects.

\section{CONCLUSION}

In this paper we first investigated the effect of using different well-known CNNs in AU detection, CNNs that vary severely in structure and computational cost. Results showed that LightCNN-9, LightCNN-29, and VGG-16 are the best-performing CNNs in AU detection, however, there is a trade off between performance and computational cost across the 3 CNNs. Second, we explored how some of the widely-used training settings (i.e. input normalization, augmentation severity, data balancing) impact the AU detection performance. Results showed that using a) sample-wise centering/normalization, b) severe augmentation, and c) oversampling improves the performance by $\sim$1-4\% for each setting. Third, we tested how the training set structure affects the AU detection performance. Results revealed that increasing the number of labelled frames for a specific number of subjects has almost no change on performance, while increasing the number of labelled subjects has a notable effect on performance. In all our analysis we used a large-scale naturalistic dataset consisting of $\sim$55K videos.




{\small
\bibliographystyle{ieee}
\bibliography{egbib}

\begin{thebibliography}{10}\itemsep=-1pt

\bibitem{antipov2015learned}
G.~Antipov, S.-A. Berrani, N.~Ruchaud, and J.-L. Dugelay.
\newblock Learned vs. hand-crafted features for pedestrian gender recognition.
\newblock In {\em Proceedings of the 23rd ACM international conference on
  Multimedia}, pages 1263--1266. ACM, 2015.

\bibitem{bishay2019schinet}
M.~Bishay, P.~Palasek, et~al.
\newblock Schinet: Automatic estimation of symptoms of schizophrenia from
  facial behaviour analysis.
\newblock {\em IEEE Transactions on Affective Computing}, 2019.

\bibitem{bishay2017fusing}
M.~Bishay and I.~Patras.
\newblock Fusing multilabel deep networks for facial action unit detection.
\newblock In {\em 2017 12th IEEE International Conference on Automatic Face \&
  Gesture Recognition (FG 2017)}, pages 681--688. IEEE, 2017.

\bibitem{chawla2009data}
N.~V. Chawla.
\newblock Data mining for imbalanced datasets: An overview.
\newblock {\em Data mining and knowledge discovery handbook}, pages 875--886,
  2009.

\bibitem{chollet2017xception}
F.~Chollet.
\newblock Xception: Deep learning with depthwise separable convolutions.
\newblock In {\em Proceedings of the IEEE conference on computer vision and
  pattern recognition}, pages 1251--1258, 2017.

\bibitem{chu2017learning}
W.-S. Chu, F.~De~la Torre, and J.~F. Cohn.
\newblock Learning spatial and temporal cues for multi-label facial action unit
  detection.
\newblock In {\em 2017 12th IEEE International Conference on Automatic Face \&
  Gesture Recognition (FG 2017)}, pages 25--32. IEEE, 2017.

\bibitem{EkmanBook97}
P.~Ekman and E.~L. Rosenberg.
\newblock {\em What the face reveals: Basic and applied studies of spontaneous
  expression using the Facial Action Coding System (FACS)}.
\newblock Oxford University Press, USA, 1997.

\bibitem{ghosh2015multi}
S.~Ghosh, E.~Laksana, S.~Scherer, and L.-P. Morency.
\newblock A multi-label convolutional neural network approach to cross-domain
  action unit detection.
\newblock In {\em ACII}, pages 609--615. IEEE, 2015.

\bibitem{gudi2015deep}
A.~Gudi, H.~E. Tasli, T.~M. Den~Uyl, and A.~Maroulis.
\newblock Deep learning based facs action unit occurrence and intensity
  estimation.
\newblock In {\em FG 2015}, volume~6, pages 1--5. IEEE, 2015.

\bibitem{hayale2019facial}
W.~Hayale, P.~Negi, and M.~Mahoor.
\newblock Facial expression recognition using deep siamese neural networks with
  a supervised loss function.
\newblock In {\em FG 2019}, pages 1--7. IEEE, 2019.

\bibitem{he2016deep}
K.~He, X.~Zhang, S.~Ren, and J.~Sun.
\newblock Deep residual learning for image recognition.
\newblock In {\em Proceedings of the IEEE conference on computer vision and
  pattern recognition}, pages 770--778, 2016.

\bibitem{huang2017densely}
G.~Huang, Z.~Liu, L.~Van Der~Maaten, and K.~Q. Weinberger.
\newblock Densely connected convolutional networks.
\newblock In {\em Proceedings of the IEEE conference on computer vision and
  pattern recognition}, pages 4700--4708, 2017.

\bibitem{jaiswal2016deep}
S.~Jaiswal and M.~Valstar.
\newblock Deep learning the dynamic appearance and shape of facial action
  units.
\newblock In {\em 2016 IEEE Winter Conference on Applications of Computer
  Vision (WACV)}, pages 1--8. IEEE, 2016.

\bibitem{jung2015deep}
H.~Jung, S.~Lee, S.~Park, I.~Lee, C.~Ahn, and J.~Kim.
\newblock Deep temporal appearance-geometry network for facial expression
  recognition.
\newblock {\em arXiv preprint arXiv:1503.01532}, 2015.

\bibitem{kahou2016emonets}
S.~E. Kahou, X.~Bouthillier, et~al.
\newblock Emonets: Multimodal deep learning approaches for emotion recognition
  in video.
\newblock {\em Journal on Multimodal User Interfaces}, 10(2):99--111, 2016.

\bibitem{krizhevsky2012imagenet}
A.~Krizhevsky, I.~Sutskever, and G.~E. Hinton.
\newblock Imagenet classification with deep convolutional neural networks.
\newblock {\em Advances in neural information processing systems},
  25:1097--1105, 2012.

\bibitem{lakshminarayana2020learning}
N.~Lakshminarayana, S.~Setlur, and V.~Govindaraju.
\newblock Learning guided attention masks for facial action unit recognition.
\newblock In {\em 2020 15th IEEE International Conference on Automatic Face and
  Gesture Recognition (FG 2020)}, pages 465--472. IEEE, 2020.

\bibitem{le2011learning}
Q.~V. Le, W.~Y. Zou, S.~Y. Yeung, and A.~Y. Ng.
\newblock Learning hierarchical invariant spatio-temporal features for action
  recognition with independent subspace analysis.
\newblock In {\em Computer Vision and Pattern Recognition (CVPR), 2011 IEEE
  Conference on}, pages 3361--3368. IEEE, 2011.

\bibitem{li2020deep}
S.~Li and W.~Deng.
\newblock Deep facial expression recognition: A survey.
\newblock {\em IEEE Transactions on Affective Computing}, 2020.

\bibitem{lu2020self}
L.~Lu, L.~Tavabi, and M.~Soleymani.
\newblock Self-supervised learning for facial action unit recognition through
  temporal consistency.
\newblock In {\em BMVC}, 2020.

\bibitem{lucey2011painful}
P.~Lucey, J.~F. Cohn, K.~M. Prkachin, P.~E. Solomon, and I.~Matthews.
\newblock Painful data: The unbc-mcmaster shoulder pain expression archive
  database.
\newblock In {\em Face and Gesture 2011}, pages 57--64. IEEE, 2011.

\bibitem{martinez2017automatic}
B.~Martinez, M.~F. Valstar, et~al.
\newblock Automatic analysis of facial actions: A survey.
\newblock {\em IEEE Transactions on Affective Computing}, 2017.

\bibitem{mavadati2013disfa}
S.~M. Mavadati, M.~H. Mahoor, et~al.
\newblock Disfa: A spontaneous facial action intensity database.
\newblock {\em IEEE Transactions on Affective Computing}, 4(2):151--160, 2013.

\bibitem{mcduff2018fed+}
D.~McDuff, M.~Amr, and R.~El~Kaliouby.
\newblock Am-fed+: An extended dataset of naturalistic facial expressions
  collected in everyday settings.
\newblock {\em IEEE Transactions on Affective Computing}, 10(1):7--17, 2018.

\bibitem{mcduff2013affectiva}
D.~McDuff, R.~Kaliouby, et~al.
\newblock Affectiva-mit facial expression dataset (am-fed): Naturalistic and
  spontaneous facial expressions collected.
\newblock In {\em Proceedings of the IEEE Conference on Computer Vision and
  Pattern Recognition Workshops}, pages 881--888, 2013.

\bibitem{nair2010rectified}
V.~Nair and G.~E. Hinton.
\newblock Rectified linear units improve restricted boltzmann machines.
\newblock In {\em Icml}, 2010.

\bibitem{niinuma2019unmasking}
K.~Niinuma, L.~A. Jeni, I.~O. Ertugrul, and J.~F. Cohn.
\newblock Unmasking the devil in the details: What works for deep facial action
  coding?
\newblock In {\em BMVC: proceedings of the British Machine Vision Conference.
  British Machine Vision Conference}, volume 2019. NIH Public Access, 2019.

\bibitem{parkhi2015deep}
O.~M. Parkhi, A.~Vedaldi, and A.~Zisserman.
\newblock Deep face recognition.
\newblock 2015.

\bibitem{sandler2018mobilenetv2}
M.~Sandler, A.~Howard, M.~Zhu, A.~Zhmoginov, and L.-C. Chen.
\newblock Mobilenetv2: Inverted residuals and linear bottlenecks.
\newblock In {\em Proceedings of the IEEE conference on computer vision and
  pattern recognition}, pages 4510--4520, 2018.

\bibitem{shao2018deep}
Z.~Shao, Z.~Liu, J.~Cai, and L.~Ma.
\newblock Deep adaptive attention for joint facial action unit detection and
  face alignment.
\newblock In {\em Proceedings of the European conference on computer vision
  (ECCV)}, pages 705--720, 2018.

\bibitem{shorten2019survey}
C.~Shorten and T.~M. Khoshgoftaar.
\newblock A survey on image data augmentation for deep learning.
\newblock {\em Journal of Big Data}, 6(1):1--48, 2019.

\bibitem{simonyan2014very}
K.~Simonyan and A.~Zisserman.
\newblock Very deep convolutional networks for large-scale image recognition.
\newblock {\em arXiv preprint arXiv:1409.1556}, 2014.

\bibitem{szegedy2016rethinking}
C.~Szegedy, V.~Vanhoucke, S.~Ioffe, J.~Shlens, and Z.~Wojna.
\newblock Rethinking the inception architecture for computer vision.
\newblock In {\em Proceedings of the IEEE conference on computer vision and
  pattern recognition}, pages 2818--2826, 2016.

\bibitem{tu2019idennet}
C.-H. Tu, C.-Y. Yang, and J.~Y.-j. Hsu.
\newblock Idennet: Identity-aware facial action unit detection.
\newblock In {\em FG 2019}, pages 1--8. IEEE, 2019.

\bibitem{vemulapalli2019compact}
R.~Vemulapalli and A.~Agarwala.
\newblock A compact embedding for facial expression similarity.
\newblock In {\em Proceedings of the IEEE conference on computer vision and
  pattern recognition}, pages 5683--5692, 2019.

\bibitem{wu2018light}
X.~Wu, R.~He, Z.~Sun, and T.~Tan.
\newblock A light cnn for deep face representation with noisy labels.
\newblock {\em IEEE Transactions on Information Forensics and Security},
  13(11):2884--2896, 2018.

\bibitem{yang2019learning}
H.~Yang and L.~Yin.
\newblock Learning temporal information from a single image for au detection.
\newblock In {\em 2019 14th IEEE International Conference on Automatic Face \&
  Gesture Recognition (FG 2019)}, pages 1--8. IEEE, 2019.

\bibitem{zhang2014bp4d}
X.~Zhang, L.~Yin, et~al.
\newblock Bp4d-spontaneous: a high-resolution spontaneous 3d dynamic facial
  expression database.
\newblock {\em Image and Vision Computing}, 32(10):692--706, 2014.

\bibitem{zhao2016deep}
K.~Zhao, W.-S. Chu, and H.~Zhang.
\newblock Deep region and multi-label learning for facial action unit
  detection.
\newblock In {\em Proceedings of the IEEE Conference on Computer Vision and
  Pattern Recognition}, pages 3391--3399, 2016.

\end{thebibliography}
}

\end{document}